\documentclass[twocolumn, switch]{article}

\usepackage{preprint}
\usepackage{placeins}
\usepackage{amsmath, amsthm, amssymb, amsfonts}
\usepackage[numbers,square]{natbib}
\usepackage[utf8]{inputenc}
\usepackage[T1]{fontenc}
\usepackage{xcolor}
\usepackage{booktabs}
\usepackage{nicefrac}
\usepackage{microtype}
\usepackage{lineno}
\usepackage{float}
\usepackage{lipsum}

\usepackage{newfloat}
\DeclareFloatingEnvironment[name={Supplementary Figure}]{suppfigure}
\usepackage{sidecap}
\sidecaptionvpos{figure}{c}

\usepackage{tikz}
\usepackage[colorlinks=true,linkcolor=blue,citecolor=blue,urlcolor=blue]{hyperref}

\definecolor{lime}{HTML}{A6CE39}
\DeclareRobustCommand{\orcidicon}{
  \begin{tikzpicture}
    \draw[lime, fill=lime] (0,0) circle [radius=0.16] 
      node[white] {{\fontfamily{qag}\selectfont \tiny ID}};
    \draw[white, fill=white] (-0.0625,0.095) circle [radius=0.007];
  \end{tikzpicture}
  \hspace{-2mm}
}
\foreach \x in {A,...,Z}{\expandafter\xdef\csname orcid\x\endcsname{\noexpand\href{https://orcid.org/\csname orcidauthor\x\endcsname}{\noexpand\orcidicon}}}

\title{LLM Encoder vs. Decoder: Robust Detection of Chinese AI-Generated Text with LoRA}

\usepackage{authblk}

\author[1,2]{Houji Jin}
\author[2]{Negin Ashrafi\orcidB{}}
\author[2]{Armin Abdollahi}
\author[1]{Wei Liu}
\author[1]{Jian Wang}
\author[1]{Ganyu Gui}
\author[2,*]{Maryam Pishgar\orcidA{}}
\author[1]{Huanghao Feng}
\affil[1]{Suzhou University of Technology, Changshu, Jiangsu, 215500, China}
\affil[2]{University of Southern California, CA, 90007, USA}
\affil[*]{Corresponding author: \texttt{pishgar@usc.edu}}
\usepackage{CJKutf8}
\usepackage{graphicx}
\usepackage{subcaption}
\usepackage{threeparttable}
\usepackage{comment}

\begin{document}
\setcounter{footnote}{0}

\twocolumn[
\maketitle
\begin{abstract}
The rapid growth of large language models (LLMs) has heightened the demand for accurate detection of AI-generated text, particularly in languages like Chinese, where subtle linguistic nuances pose significant challenges to current methods. In this study, we conduct a systematic comparison of encoder-based Transformers (Chinese BERT-large and RoBERTa-wwm-ext-large), a decoder-only LLM (Alibaba’s Qwen2.5-7B/DeepSeek-R1-Distill-Qwen-7B fine-tuned via Low-Rank Adaptation (LoRA)), and a FastText baseline using the publicly available dataset from the NLPCC 2025 Chinese AI-Generated Text Detection Task. Encoder models were fine-tuned using a novel prompt-based masked language modeling approach, while Qwen2.5-7B was adapted for classification with an instruction-format input and a lightweight classification head trained via LoRA. Experiments reveal that, although encoder models nearly memorize training data, they suffer significant performance degradation under distribution shifts (RoBERTa: 76.3\% test accuracy; BERT: 79.3\%). FastText demonstrates surprising lexical robustness (83.5\% accuracy) yet lacks deeper semantic understanding. In contrast, the LoRA-adapted Qwen2.5-7B achieves 95.94\% test accuracy with balanced precision–recall metrics, indicating superior generalization and resilience to dataset-specific artifacts. These findings underscore the efficacy of decoder-based LLMs with parameter-efficient fine-tuning for robust Chinese AI-generated text detection. Future work will explore next-generation Qwen3 models, distilled variants, and ensemble strategies to further enhance cross-domain robustness.
\end{abstract}
\vspace{0.35cm}
]
\begin{CJK*}{UTF8}{gbsn}

\section{Introduction}
\label{sec:introduction}

Large Language Models (LLMs) can now generate text that closely mimics human writing, raising concerns about misuse and the difficulty of distinguishing AI-generated content from human-written text \citet{ref1, ref2, ref3}. In fact, studies have shown that humans perform only slightly better than random chance when trying to detect AI-generated text \citet{ref4, ref5}. This has motivated extensive research into LLM-generated text detection, which aims to discern if a given text was produced by an AI or a human – essentially a binary classification problem \citet{ref2, ref3, ref6}.

This challenge has taken on growing importance across domains such as education, journalism, and scientific publishing, where trust and authorship are critical. In parallel, the detection of AI-generated Chinese text presents additional complexity due to its character-level structure, absence of explicit word boundaries, and rich semantic ambiguity \citet{ref12, ref13}. Despite this, relatively few studies have rigorously evaluated model architectures tailored for Chinese AI-text detection \citet{ref14, ref15}.

The NLPCC 2025 Shared Task 1, LLM-Generated Text Detection, introduced a publicly available dataset of Chinese texts annotated as either human-written or AI-generated. In this study, we use that dataset to investigate how different model architectures—encoder-based Transformers (such as BERT \citet{ref7} and RoBERTa \citet{ref8}) versus decoder-based LLMs (such as the GPT-style Qwen family \citet{ref9, ref10}), perform on the task of Chinese AI-generated text detection.

In this paper, we compare representative encoder and decoder models on the binary classification task of detecting AI-generated Chinese text. We fine-tune two prevalent encoder models (Chinese BERT and RoBERTa \citet{ref11}) and strong decoder LLMs (Alibaba’s Qwen-7B and DeepSeek-R1-Distill-Qwen-7B \citet{ref10, ref24}) on the same dataset, and also include a classical linear baseline (FastText \citet{ref22}) for reference. We then evaluate their precision, recall, and F1 performance on training, development, and test sets. This comparative study of encoder versus decoder Transformer architectures highlights the relative strengths of each approach and provides practical insights for building robust detectors of Chinese AI-generated text.

The key contributions of this study are threefold. First, we provide the first direct comparison between encoder and decoder architectures for Chinese AI-generated text detection under identical conditions. Second, we introduce a prompt-based masked language modeling approach for encoders that improves out-of-domain generalization. Third, we demonstrate that a LoRA-adapted decoder (Qwen2.5-7B) achieves state-of-the-art robustness, substantially outperforming both strong encoder baselines and classical lexical models.

To guide the reader through the remainder of this paper, Section~\ref{sec:related} reviews prior work on AI-generated text detection with a focus on Chinese. Section~\ref{sec:model} presents the selected models, including encoder-based Transformers, decoder-only LLMs, and a lexical baseline. Section~\ref{sec:metho} outlines the training and evaluation methodology. Section~\ref{sec:results} reports the experimental findings and provides comparative analysis. Finally, Section~\ref{sec:conc} concludes with implications and directions for future research.

\section{Related Work}
\label{sec:related}
Early efforts on AI-generated text detection mostly employed encoder-based classifiers. For instance, Petropoulos et al. fine-tuned a RoBERTa model to classify whether a given text (in English or Chinese) was written by a human or by a model \citet{ref18}. Such BERT/RoBERTa-based detectors achieved strong in-domain accuracy, though they risk overfitting to the writing style of seen AI models and may struggle on out-of-domain data. On the other hand, researchers have explored zero-shot detection methods that do not require training data. One prominent example is DetectGPT \citet{ref3}, which examines the probability curvature of a text under a language model; AI-generated text tends to have lower likelihood than paraphrases of the same text. While zero-shot methods like DetectGPT and others based on perplexity or burstiness can flag AI-generated text without supervised training, they often require long inputs to be reliable and their accuracy drops on shorter texts, which is a limitation for sentence-level detection \citet{ref16}.

Another line of work uses watermarking: imperceptibly imprinting a signal in AI-generated text so it can later be identified. Kirchenbauer et al. introduced a cryptographic watermarking method for LLM outputs \citet{ref17}. However, watermark approaches have downsides: they can slightly degrade text quality, and adversaries may remove or obfuscate the watermark, so these methods are not foolproof \citet{ref20}. Furthermore, watermarking requires control over the text generation process, which is not applicable if we need to detect arbitrary AI text ex-post \citet{ref19}.

More recently, the emergence of large instruct-tuned models has opened a new avenue: using the LLMs themselves as detectors \citet{ref14}. The intuition is that an LLM possesses a great deal of knowledge from pre-training, potentially including subtle cues of AI-generated text. Similar strategies leveraging pretraining and teacher–student architectures have also shown strong performance in other domains \citet{sepsis}. Instruction tuning an LLM on the detection task can align it to produce the desired classification behavior. For example, the LLM-Detector system fine-tuned open-source LLMs (7B–13B parameters) with detection instructions and achieved promising results, outperforming earlier methods on both in-domain and cross-domain detection \citet{ref14}. These advances suggest that decoder-based models, despite being generative, can be adapted into powerful classifiers for AI-generated text.

Overall, previous research has demonstrated the effectiveness of encoder-based models for detection, and more recent evidence shows that instruction-tuned decoder LLMs can match or even surpass their performance. However, there has been little direct comparison between encoder and decoder architectures on the same data to quantify their differences. Our work addresses this gap by evaluating both approaches on Chinese AI-generated text detection under identical conditions.

\section{Model Selection and Overview}
\label{sec:model}

\subsection{Chinese RoBERTa-wwm-ext-large}
This model is a transformer encoder based on the BERT-large architecture tailored for Chinese. It contains 24 layers of bidirectional Transformers with a hidden size of 1024 and 16 self-attention heads. The model has on the order of 330 million parameters (similar to BERT-large). It was pretrained with whole-word masking on large Chinese corpora: starting from the Chinese Wikipedia (roughly 0.4 billion words) and then extended with an additional 5.4 billion words of text from encyclopedias, news articles, and QA forums. This extended corpus is over ten times larger than Wikipedia and provides broader coverage of vocabulary and styles. RoBERTa-wwm-ext-large was trained with a masked language modeling objective (like BERT) but without the next-sentence prediction task (following the RoBERTa training protocol). The whole-word masking means if a Chinese word (comprised of multiple characters) is selected for masking, all its characters are masked together, helping the model learn complete word representations. In summary, Chinese RoBERTa-wwm-ext-large offers a powerful encoder with deep bidirectional context and was exposed to a wide range of formal and informal Chinese text during pretraining.

\subsection{Chinese BERT-large}
This model shares the same architecture as RoBERTa-large – 24 transformer encoder layers, hidden size 1024, 16 attention heads – but was one of the earlier Chinese pre-trained models using BERT’s original training procedure. Cui et al. trained a Chinese BERT-large with whole-word masking as well, initializing from the official BERT-base (Chinese) vocabulary and weights for consistency \citet{ref21}. Because the original BERT release included only a Chinese base model, the authors trained the large model from scratch on Chinese data. The corpora used are similar (Wikipedia plus extended data) to RoBERTa-wwm-ext, but BERT-large was trained with the next sentence prediction (NSP) objective in addition to masked language modeling. In practice, Chinese BERT-large and RoBERTa-large have nearly identical model structure, and both leverage whole-word masking to better capture Chinese word-level semantics. We include both to compare any differences due to training regimen: BERT-large (Chinese) represents a classic encoder with NSP, while RoBERTa-wwm-ext-large represents an NSP-free encoder trained on even more data.

\subsection{FastText classifier}
As a lightweight baseline, we use FastText for Chinese text classification. FastText is a shallow neural model that learns word embeddings jointly with the classifier from scratch, without any pretrained vectors. The input texts are preprocessed using jieba.lcut for segmentation, and the model generates token-level embeddings directly during training. FastText efficiently captures lexical patterns through character-level and subword n-grams, and averages token embeddings into a fixed-size vector, subsequently classified by a linear softmax layer. Despite its simplicity (embedding dimension defaulting to 100, adjustable if explicitly set, no attention mechanisms or transformer layers), FastText is parameter-efficient and serves as an efficient baseline to compare against more sophisticated transformer-based models. FastText lacks transformer layers or attention mechanisms, consisting solely of embedding lookup, averaging, and a linear output layer, thus making it highly parameter-efficient and computationally inexpensive. Its strength lies in efficiently capturing surface-level lexical patterns and n-gram-level features. We included FastText as a baseline specifically to gauge how effectively simpler, non-contextual embedding-based methods can discriminate between human-written and AI-generated Chinese texts, thus providing a valuable contrast to the more complex and computationally intensive transformer-based models. However, due to its lack of contextual modeling capabilities, FastText cannot fully exploit deep semantic or syntactic features present in the text.

\subsection{Qwen2.5-7B}
Qwen2.5-7B is a large-scale decoder model from Alibaba Cloud’s Qwen series, with about 7.6 billion parameters. It follows a transformer decoder-only architecture (similar to GPT) with 28 layers, a hidden size of 3584, and a grouped-query attention design (GQA) using 28 query heads and 4 key/value heads. This GQA configuration reduces memory by sharing keys/values across fewer heads while keeping many query heads for expressiveness. Qwen2.5 uses rotary positional embeddings (RoPE) to handle very long context lengths (up to 131 k tokens), an enhanced SwiGLU activation \citet{ref23} in feed-forward layers, and RMSNorm normalization. The model was pretrained on an extremely large and diverse corpus—on the order of 2.4 trillion tokens—including web pages, news, code, and conversations in both Chinese and English. This massive pretraining gives Qwen2.5 broad knowledge and the ability to understand complex language patterns. The specific variant we use is the base 7 B model (not instruction-tuned by default) which we fine-tune for the detection task. Chinese Qwen2.5-7B is essentially a powerful generative model (similar scale to GPT-3’s smaller variants) that we adapt as a classifier. Its architecture is decoder-only (unidirectional attention) but with enough layers and data to capture rich linguistic features, potentially useful for distinguishing human vs. AI text.

\subsection{DeepSeek-R1-Distill-Qwen-7B}

This model is a specialized derivative of Qwen, produced by the DeepSeek project. DeepSeek-R1 is a large reasoning-centric LLM (comparable to OpenAI’s GPT-4 variants in some benchmarks) which the team distilled down into smaller models for public use. DeepSeek-R1-Distill-Qwen-7B is essentially the Qwen2.5-7B architecture fine-tuned on DeepSeek’s proprietary reasoning data. It retains the same architecture as Qwen2.5-7B—28 decoder layers, hidden size 3584, 28 query/4 key heads—but has been instruction-tuned or optimized to emulate the reasoning behavior of the much larger DeepSeek-R1 model. According to the model card, the distillation used ~800 k curated examples that force the 7B model to mimic the larger model’s responses. The result is a 7B decoder model exceptionally good at reasoning and following instructions, outperforming many other open models of similar size. We include this model to answer whether reasoning-optimized distillation gives an edge in detecting AI-generated text. Its underlying pretraining corpora mirror Qwen2.5-7B; only the distillation stage adds specialized reasoning data.

\section{Methodology}
\label{sec:metho}
\begin{figure*}[pht]
    \centering
    \includegraphics[width=0.8\textwidth]{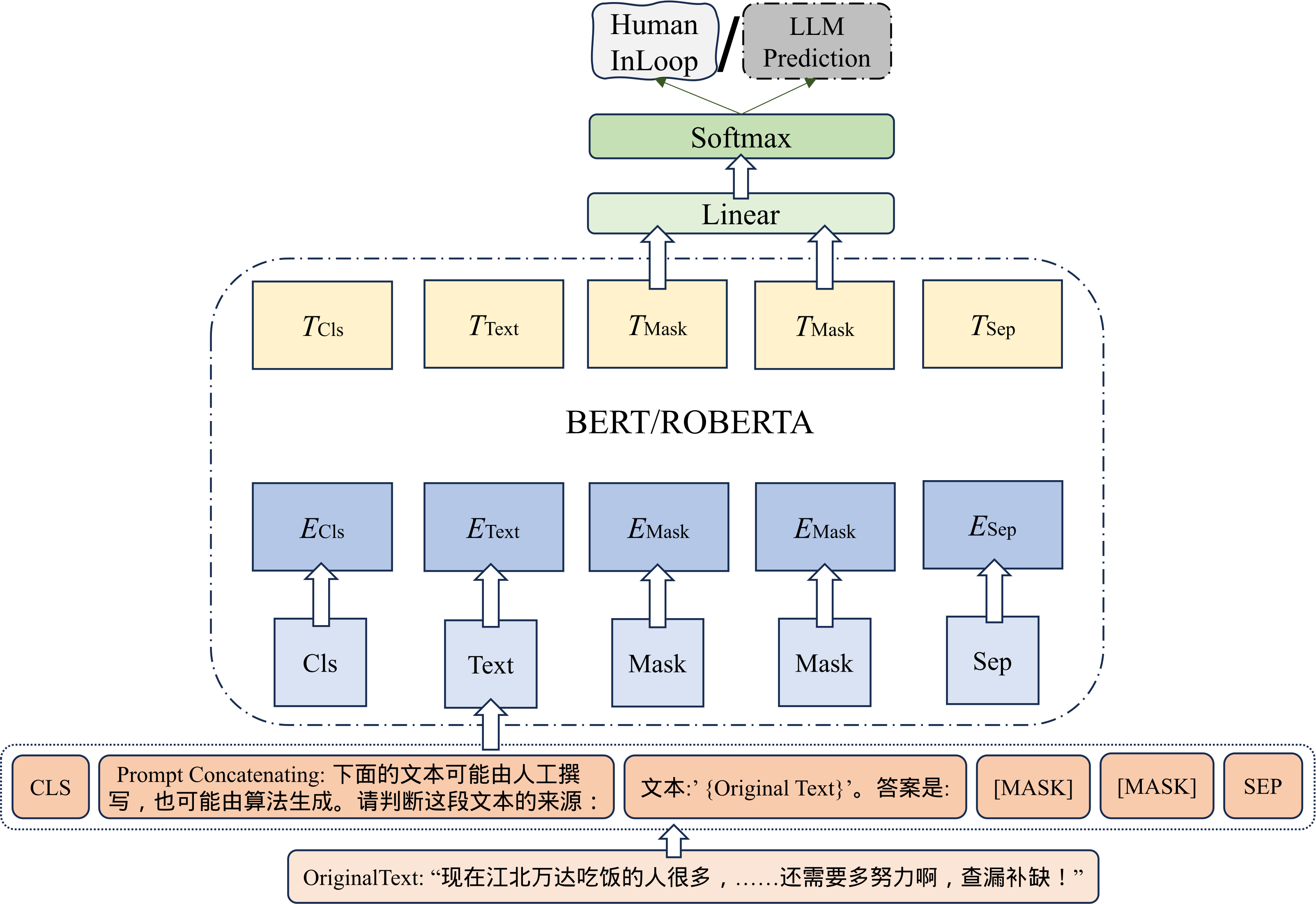}
    \caption{Encoder based model training logic and prompt design architecture.}
    \label{fig:encoder}
\end{figure*}

\begin{figure*}[ht]
    \centering
    \includegraphics[width=0.6\textwidth]{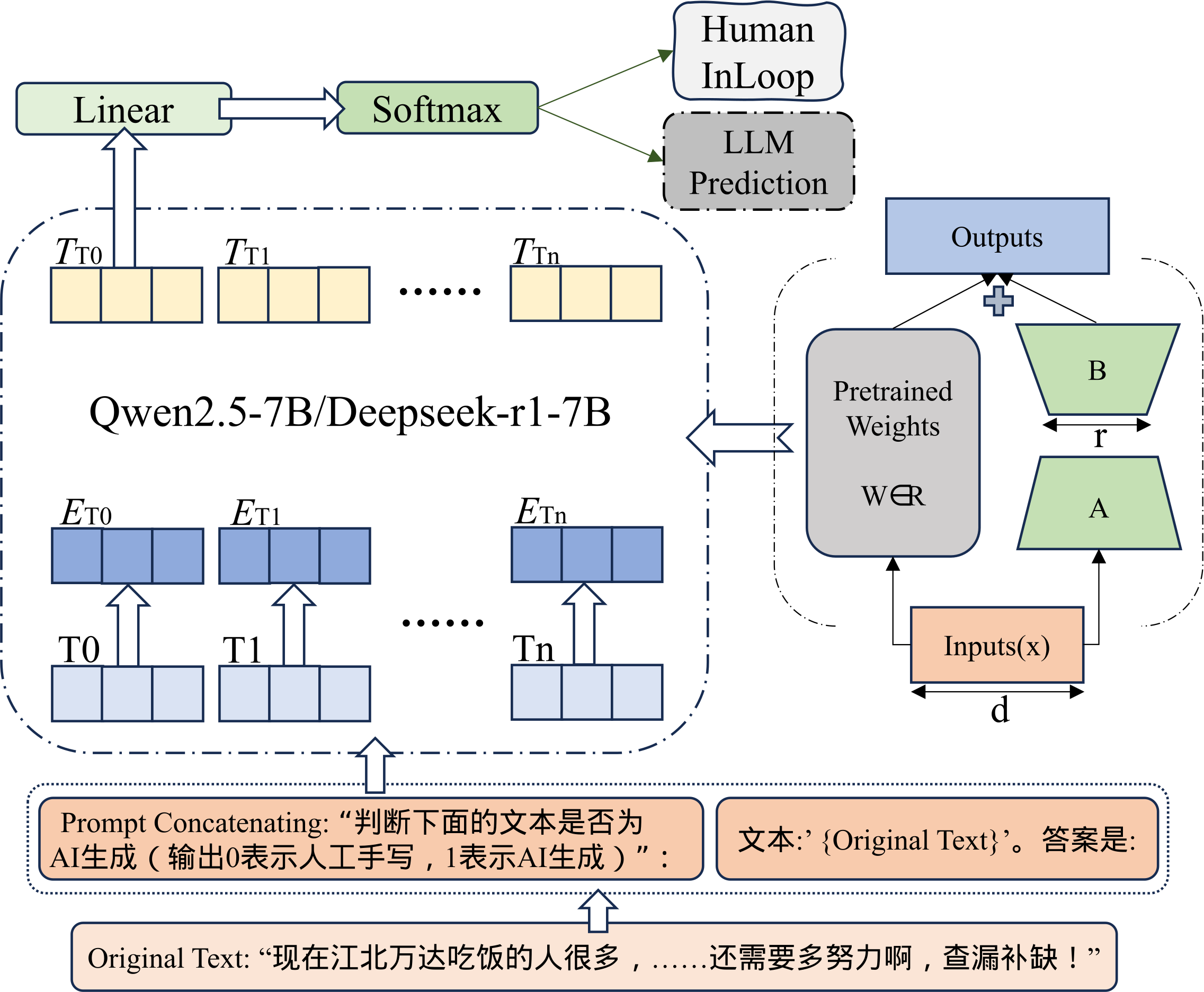}
    \caption{Decoder based model architecture comprises standard components, token embedding with positional encoding, masked multi-head self-attention, and feed-forward layers, and is augmented by specialized modules labeled “E” and “T.”}
    \label{fig:decoder}
\end{figure*}

We cast LLM-generated text detection as a binary classification problem in which each Chinese sentence (or short paragraph) must be labelled as \emph{human-written} or \emph{AI-generated}. Three model families are examined under an identical train–development–test split from the publicly available NLPCC-2025 Shared Task 1 dataset (\url{https://github.com/NLP2CT/NLPCC-2025-Task1}): two pre-trained transformer encoders (Chinese RoBERTa-wwm-ext-large and Chinese BERT-large), two pre-trained transformer decoders (Qwen2.5-7B and the reasoning-distilled DeepSeek-R1-7B), and a lightweight lexical baseline (fastText). All neural models are fine-tuned on the training split and selected by macro-$F_{1}$ on the development set; final metrics are reported on the held-out test set. The overall encoder pipeline and prompt template are illustrated in \hyperref[fig:encoder]{Figure~\ref*{fig:encoder}}, whereas the decoder adaptation we employ is depicted in \hyperref[fig:decoder]{Figure~\ref*{fig:decoder}}.

\setlength{\floatsep}{5pt plus 2pt minus 2pt}
\setlength{\textfloatsep}{5pt plus 2pt minus 2pt}
\setlength{\intextsep}{5pt plus 2pt minus 2pt}
For the encoders we abandon the conventional ``\texttt{[CLS]} + soft-max'' head in favour of a prompt-based masked-token formulation that exploits the models' original masked-language-model pre-training.  Each input text $x$ is embedded in the following Chinese template:
\[
{\small
\begin{gathered}
\texttt{[CLS]}\ \text{下面的文本可能由人工撰写,也可能由算法生成.} \\
\text{请判断这段文本的来源:文本:'}x\text{'.答案是:}\ \texttt{[MASK][MASK]}\text{.[SEP]}
\end{gathered}}
\]
The consecutive \texttt{[MASK]} positions correspond to the bi-character words ``人工'' (human) and ``算法'' (AI).  During fine-tuning we minimize cross-entropy loss only on those masked tokens, encouraging the model to predict the correct verbaliser pair given the full bi-directional context. All encoder parameters remain trainable; we optimize with AdamW, a linearly decayed learning rate and early stopping on development-set $F_{1}$.  Empirically, this constrained Masked Language Modeling (MLM) objective regularizes the encoders and delivers better out-of-domain generalization than a randomly-initialized classification layer that attends solely to the \texttt{[CLS]} embedding.

Decoder-only Transformers pose a different challenge because their vocabularies contain no \texttt{[CLS]} token and their self-attention is uni-directional.  We therefore prefix each input with a concise Chinese instruction---``判断下面的文本是否为 AI 生成（0 = 人写，1 = AI）：''---and feed the resulting sequence into Qwen2.5-7B or DeepSeek-R1-7B. Qwen2.5-7B model has not been instruction fine-tuned, in order to standardize the training logic of the decoder, the model was not fine-tuned with instructions; instead, the final hidden state of the [CLS] token was directly employed for classification tasks. After processing the entire sequence left-to-right, the model's last hidden state for the \emph{first} token, denoted $h_{0}^{(L)}$, is treated as a sentence representation; a small linear layer maps $h_{0}^{(L)}$ to two logits whose soft-max yields class probabilities.  To avoid updating all $7$ billion backbone parameters, we apply LoRA, inserting trainable rank-$r$ matrices into every projection while freezing the original weights.  The classification head and LoRA parameters are trained jointly with cross-entropy loss. DeepSeek-R1-7B follows the same procedure but begins with weights distilled from a much larger DeepSeek-R1 teacher, allowing us to test whether reasoning-oriented knowledge transfer aids detection.  This adaptation, including the use of $h_{0}^{(L)}$ in lieu of a \texttt{[CLS]} token, is visualised in \hyperref[fig:decoder]{Figure~\ref*{fig:decoder}}.


\setlength{\floatsep}{5pt plus 2pt minus 2pt}
\setlength{\textfloatsep}{5pt plus 2pt minus 2pt}
\setlength{\intextsep}{5pt plus 2pt minus 2pt}
The FastText baseline is trained on the identical corpus after segmentation with \texttt{jieba.lcut}.  It learns bag-of-subword embeddings of dimension which are averaged and passed through a linear soft-max classifier. We set wordngrams to 2, and turn on the Autotune mode to train the model. The final hyperparam are lr=0.05, buch size= 1490212. Although it lacks contextual modelling, FastText still establishes a valuable lower bound. When it achieves high accuracy, shallow lexical cues suffice; however, if its performance falls behind, that gap demonstrates the necessity of the deep semantic representations that transformers capture.

\section{Experimental Results and Analysis}
\label{sec:results}

\subsection{Experimental Setup}

We fine–tuned every model on the official training split of the NLPCC--2025 Task~1 corpus and selected checkpoints by macro-$F_{1}$ on the development set. The corpus comprises short Chinese texts spanning multiple domains, each annotated as \emph{human-written} or \emph{AI-generated}; both the dev and test partitions are approximately class-balanced. Fine-tuning for the encoder models (RoBERTa-wwm-ext-large and BERT-large) followed the masked-prompt recipe described earlier; for the decoder model (Qwen2.5-7B) we employed LoRA with rank 4,8, and 16 and a linear classification head on the first-token representation. FastText was trained on the same tokenized data (wordngrams to 2, and turn on the Autotune mode to train the model, lr=0.05, buch size= 1490212) and optimized with SGD. For training  encoder models (RoBERTa-wwm-ext-large and BERT-large) we use the AdamW optimizer with an lz"weight-decay" penalty of 0.01 applied to all model parameters except biases and LayerNorm parameters, which are excluded from decay. This provides stronger regularization on the bulk of the network weights while avoiding over-penalizing normalization layers and biases. In addition, to prevent overfitting we implement an early-stopping criterion: if the validation loss fails to improve for a fixed number of update steps (as defined by the patience parameter), training is terminated and the best-performing model on the validation set is retained, an approach also widely adopted in other machine learning applications \citet{stroke}.

\subsection{Results and Discussion}

\hyperref[tab:results]{Table~\ref*{tab:results}} summarizes performance on the train, development, and test sets. On the training split, all models memorize the data almost perfectly: RoBERTa attains $99.65\%$ accuracy, BERT $98.72\%$, and even the shallow FastText baseline reaches $99.89\%$. These figures indicate strong separability in the training distribution, with FastText’s $F_{1}$ scores ($0.9977$ for human, $0.9992$ for AI) suggesting that lexical and $n$-gram cues alone suffice for near-perfect fitting.

\begin{table*}[ht]
\centering
\renewcommand{\arraystretch}{1.4}
\caption{Performance evaluation on the NLPCC--2025 Task~1 corpus. Precision (P), Recall (R), and $F_{1}$ are reported per class; Acc.\ denotes overall accuracy.}
\label{tab:results}
\small
\begin{threeparttable}
\setlength{\tabcolsep}{7pt}
\begin{tabular}{llccccccc}
\toprule
\textbf{Model} & \textbf{Split} &
$\mathbf{P_{H}}$ & $\mathbf{R_{H}}$ & $\mathbf{F1_{H}}$ &
$\mathbf{P_{AI}}$ & $\mathbf{R_{AI}}$ & $\mathbf{F1_{AI}}$ &
\textbf{Acc.} \\
\midrule

\textbf{RoBERTa} & Train & 0.9872 & 0.9989 & 0.9930 & 0.9996 & 0.9957 & 0.9976 & 0.9965 \\
                 & Dev   & 0.9425 & 0.9236 & 0.9330 & 0.9512 & 0.9635 & 0.9573 & 0.9479 \\
                 & Test  & 0.6962 & 0.9336 & 0.7976 & 0.8993 & 0.5925 & 0.7143 & 0.7631 \\
\cmidrule(lr){1-9}

\textbf{BERT}    & Train & 0.9671 & 0.9823 & 0.9746 & 0.9941 & 0.9888 & 0.9915 & 0.9872 \\
                 & Dev   & 0.9444 & 0.8036 & 0.8684 & 0.8841 & 0.9694 & 0.9248 & 0.9043 \\
                 & Test  & 0.7620 & 0.8531 & 0.8050 & 0.8331 & 0.7334 & 0.7801 & 0.7933 \\
\cmidrule(lr){1-9}

\textbf{fastText} & Train & 0.9963 & 0.9991 & 0.9977 & 0.9997 & 0.9988 & 0.9992 & 0.9989 \\
                  & Dev   & 0.9794 & 0.8227 & 0.8943 & 0.8961 & 0.9888 & 0.9402 & 0.9236 \\
                  & Test  & 0.8350 & 0.8351 & 0.8350 & 0.8350 & 0.8349 & 0.8350 & 0.8350 \\
\cmidrule(lr){1-9}


\end{tabular}
\end{threeparttable}
\end{table*}

Generalization gaps emerge on the development set. RoBERTa remains the most reliable encoder (accuracy $94.79\%$) although its human-class precision and recall fall by roughly five points.  BERT drops to $90.43\%$ accuracy and shows a pronounced recall deficit for human texts ($0.8036$), implying a bias towards predicting the AI class. FastText, surprisingly, holds $92.36\%$ accuracy, outperforming BERT despite its lack of contextual modelling; its error profile, however, is skewed, with high human precision ($0.9794$) but low recall ($0.8227$), indicating over-reliance on surface markers.

The held-out test split exposes much sharper contrasts. RoBERTa’s accuracy plunges to $76.31\%$, driven by a collapse in human-class precision ($0.6962$) and AI-class recall ($0.5925$); clearly, it overfit to training idiosyncrasies.  BERT fares slightly better at $79.33\%$ with a more balanced confusion pattern, yet still loses almost twenty points relative to its training score.  FastText now surpasses both encoders at $83.50\%$, maintaining near-equal precision and recall for both classes; evidently, the unseen test texts preserve lexical regularities that FastText can exploit even when Transformers falter.

\hyperref[tab:qwen_deepseek_classification_reports]{Table~\ref*{tab:qwen_deepseek_classification_reports}} explains the class-wise precision, recall, F1-score and overall accuracy for Qwen2.5-7B and DeepSeek-R1-7B when fine-tuned with LoRA at ranks $r=4$, $8$, and $16$. Across every setting, Qwen2.5-7B consistently outperforms DeepSeek-R1-7B. For instance, at $r=4$, Qwen2.5-7B achieves 94.31\% accuracy (AI F1=0.9460, human F1=0.9398) versus 90.79\% for DeepSeek-R1-7B (AI F1=0.9155, human F1=0.8988). Raising the rank to $r=8$ yields modest gains for both models, but at $r=16$ Qwen2.5-7B reaches its peak 95.94\% accuracy with balanced AI (F1=0.9609) and human (F1=0.9577) scores, while DeepSeek-R1-7B tops out at 92.93\% accuracy (AI F1=0.9339, human F1=0.9240). These results underscore that the base Qwen2.5-7B decoder, even when lightly adapted via LoRA, delivers superior generalization and robustness under distribution shift, whereas the reasoning-distilled DeepSeek-R1-7B, despite its targeted supervision, remains somewhat behind in overall detection performance. We hypothesize that models without instruction fine-tuning might better capture the inherent logic of model-generated text; however, this remains speculative. Reviewing the documentation on DeepSeek distillation, it appears that while reinforcement post-training was applied without supervised fine-tuning, this does not invalidate our conjecture. 

\begin{table*}[htbp]
    \centering
    \renewcommand{\arraystretch}{1.2} 
    \caption{Qwen2.5-7B and DeepSeek-R1-7B Classification Reports on test set}
    \label{tab:qwen_deepseek_classification_reports}
    \begin{tabular}{lcccccccccc}
        \toprule
        & \multicolumn{3}{c}{R4} & \multicolumn{3}{c}{R8} & \multicolumn{3}{c}{R16} \\
        \cmidrule(lr){2-4} \cmidrule(lr){5-7} \cmidrule(l){8-10}
        & Prec & Rec & F1 & Prec & Rec & F1 & Prec & Rec & F1 \\
        \midrule
        \multicolumn{10}{l}{(Qwen2.5-7B)} \\
        \midrule
        AI           & 0.8995 & 0.9976 & \textbf{0.9460} & 0.8898 & 0.9991 & \textbf{0.9413} & 0.9269 & 0.9975 & \textbf{0.9609} \\
        Human        & \textbf{0.9973} & 0.8885 & 0.9398 & 0.9990 & 0.8762 & 0.9335 & 0.9972 & 0.9213 & 0.9577 \\
        \midrule
        Accuracy     & 0.9431 & 0.9431 & 0.9431 & 0.9376 & 0.9376 & 0.9376 & 0.9594 & 0.9594 & 0.9594 \\
        \midrule
        Macro Avg    & 0.9484 & 0.9431 & 0.9429 & 0.9444 & 0.9376 & 0.9374 & 0.9621 & 0.9594 & 0.9593 \\
        Weighted Avg & 0.9484 & 0.9431 & 0.9429 & 0.9443 & 0.9376 & 0.9374 & 0.9620 & 0.9594 & 0.9593 \\
        \midrule
        \multicolumn{10}{l}{(DeepSeek-R1-7B)} \\
        \midrule
        AI           & 0.8458 & \textbf{0.9978} & 0.9155 & 0.8195 & \textbf{0.9998} & 0.9008 & 0.8772 & \textbf{0.9984} & 0.9339 \\
        Human        & \textbf{0.9973} & 0.8180 & 0.8988 & \textbf{0.9998} & 0.7798 & 0.8762 & \textbf{0.9981} & 0.8602 & 0.9240 \\
        \midrule
        Accuracy     & 0.9079 & 0.9079 & 0.9079 & 0.8898 & 0.8898 & 0.8898 & 0.9293 & 0.9293 & 0.9293 \\
        \midrule
        Macro Avg    & 0.9216 & 0.9079 & 0.9072 & 0.9097 & 0.8898 & 0.8885 & 0.9376 & 0.9293 & 0.9289 \\
        Weighted Avg & 0.9215 & 0.9079 & 0.9072 & 0.9096 & 0.8898 & 0.8885 & 0.9376 & 0.9293 & 0.9289 \\
        \bottomrule
    \end{tabular}
\end{table*}

Across the models, the decoder-based Qwen2.5-7B, fine-tuned via LoRA with r=16, achieves the best robustness with $95.94\%$ test accuracy and balanced metrics (human $P=0.9972$, $R=0.9213$; AI $P=0.9269$, $R=0.9975$). Its large-scale generative pre-training furnishes richer linguistic representations that survive distribution shift, while LoRA regularization prevents overfitting that afflicted the encoders. These findings confirm that a lightly adapted decoder LLM can outperform smaller encoders and even a strong lexical baseline on out-of-domain data.  In practical deployments where the provenance of Chinese texts is unconstrained, we therefore recommend decoder-style detectors; encoder models remain attractive for in-domain settings provided they are augmented with stronger regularization or data augmentation, whereas fastText offers a computationally frugal first-pass filter when resources are scarce.

\begin{figure}[ht]
  \centering
  \begin{subfigure}[b]{0.15\textwidth}
    \includegraphics[width=\textwidth]{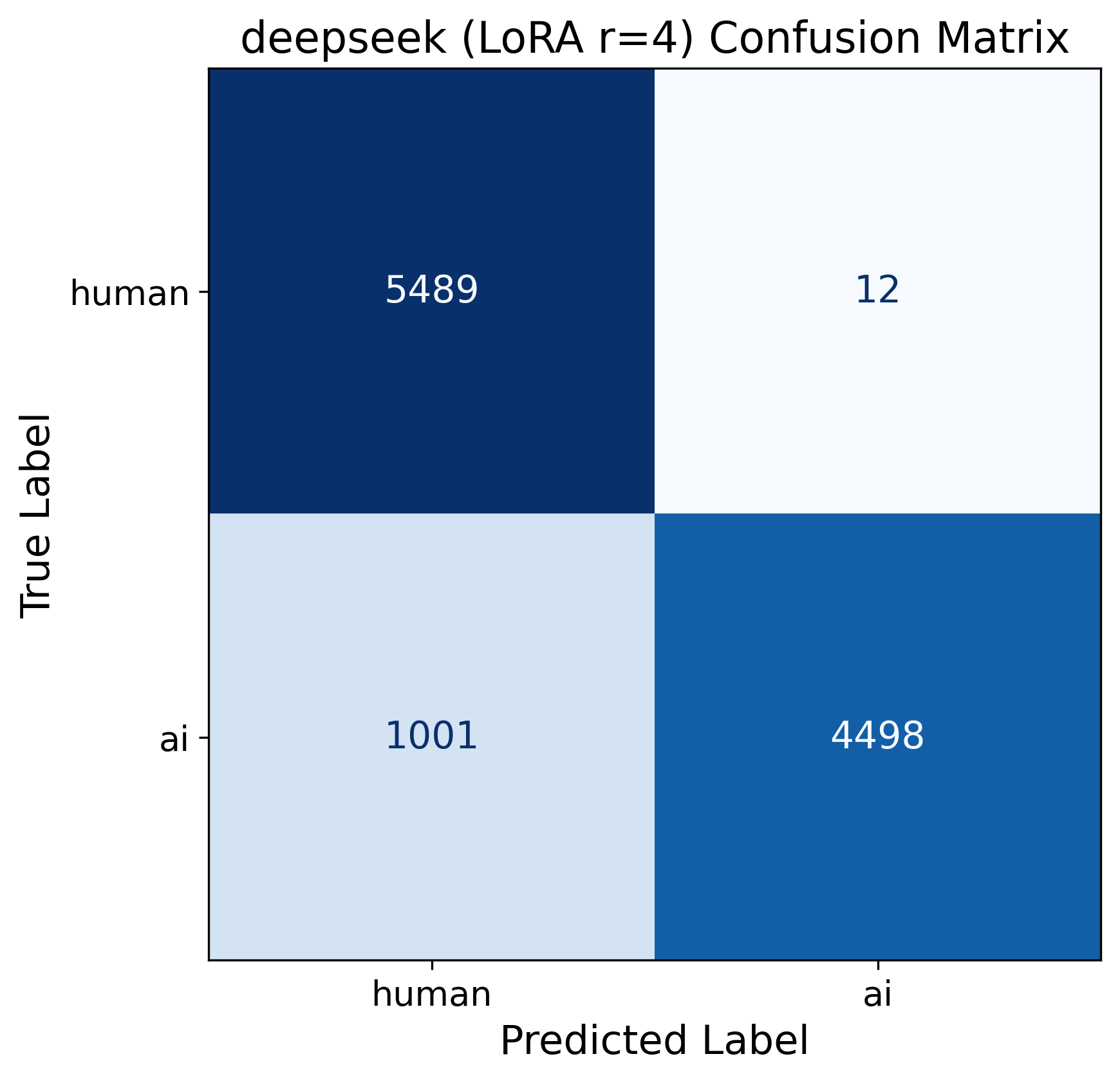}
    \caption{$r=4$}
    \label{fig:deepseek_r4_conf}
  \end{subfigure}%
  \begin{subfigure}[b]{0.15\textwidth}
    \includegraphics[width=\textwidth]{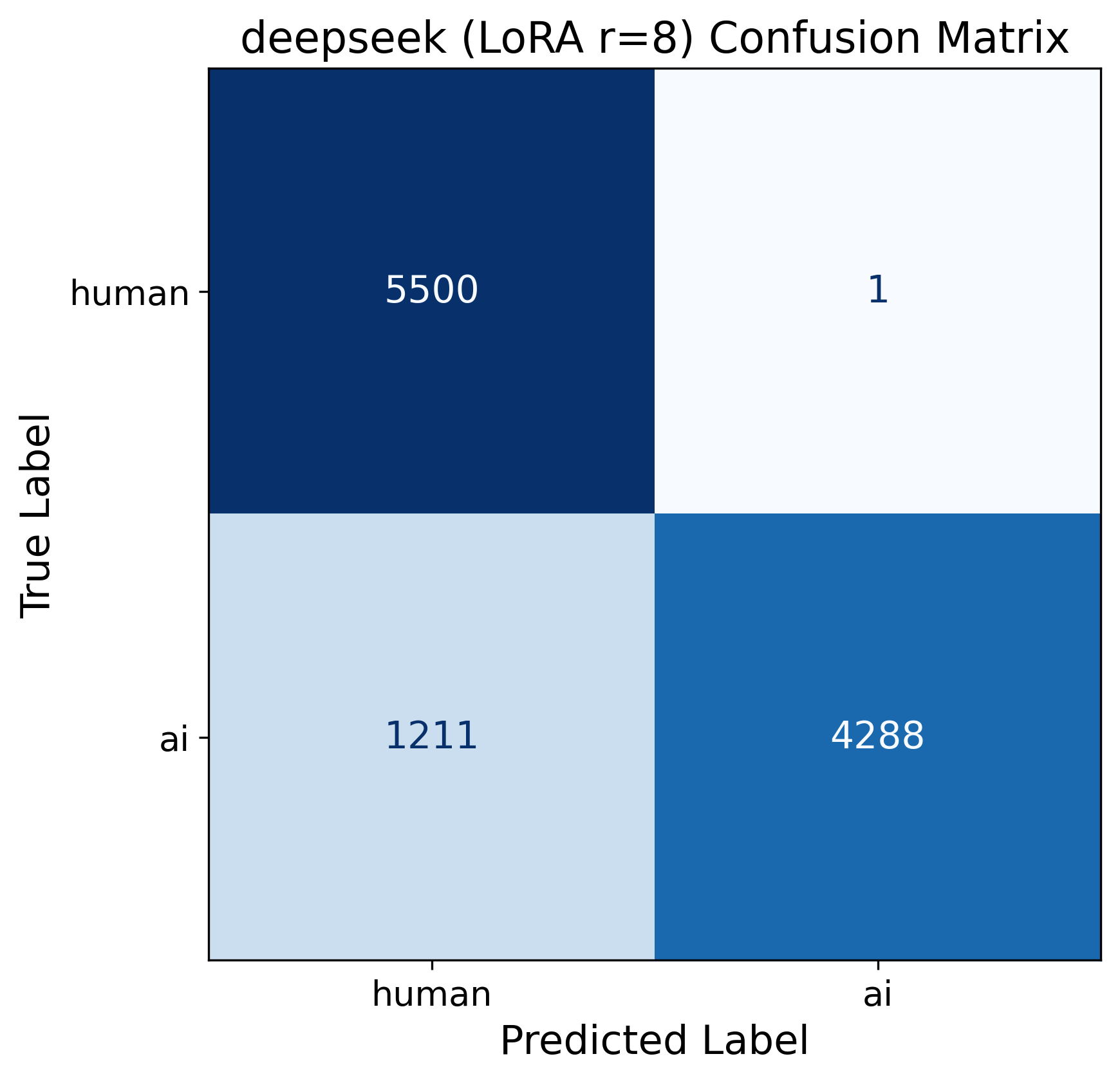}
    \caption{$r=8$}
    \label{fig:deepseek_r8_conf}
  \end{subfigure}%
  \begin{subfigure}[b]{0.15\textwidth}
    \includegraphics[width=\textwidth]{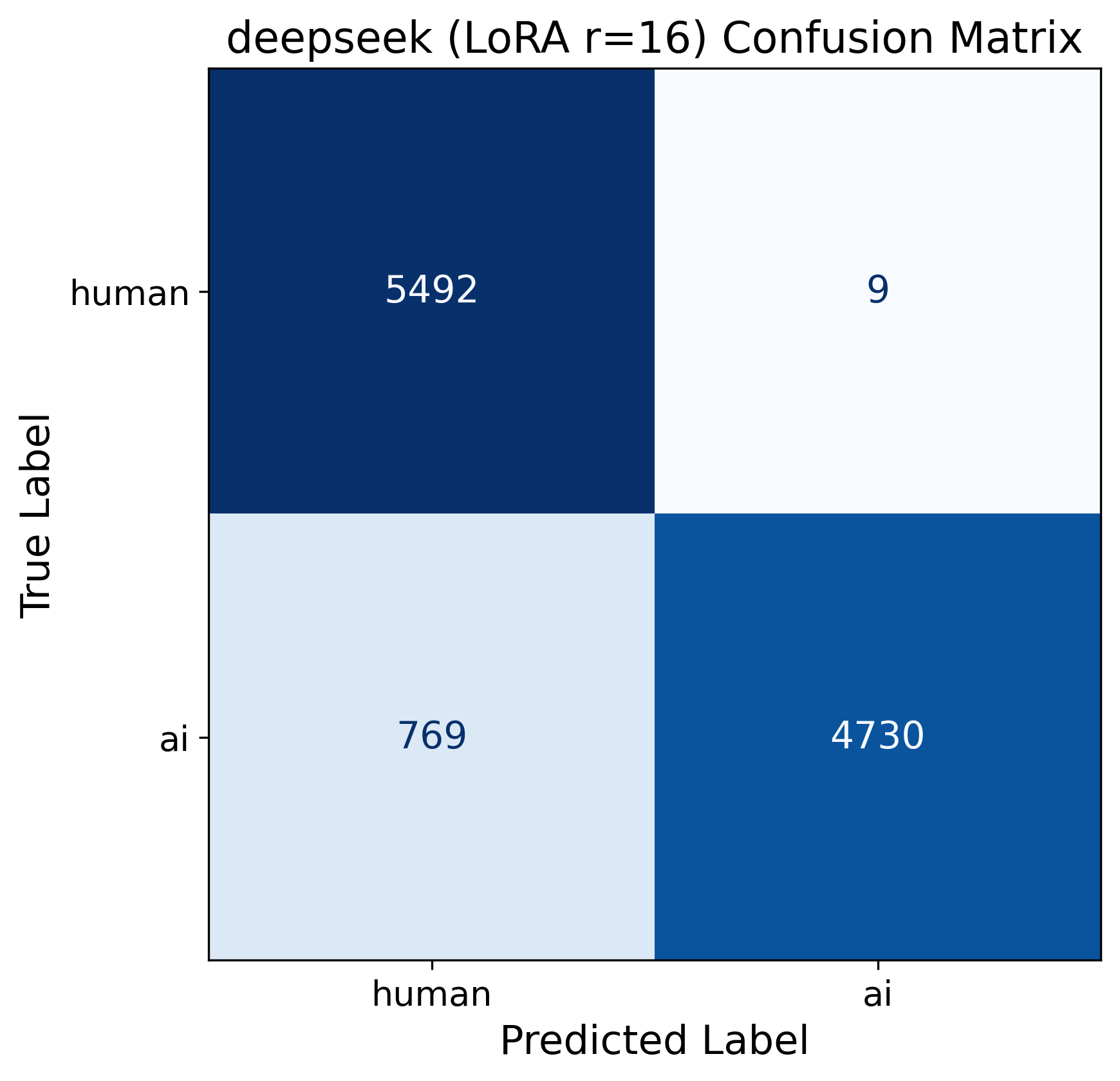}
    \caption{$r=16$}
    \label{fig:deepseek_r16_conf}
  \end{subfigure}
  \caption{DeepSeek confusion matrices for $r=4,8,16$.}
  \label{fig:deepseek_confusion}
\end{figure}

\hyperref[fig:deepseek_confusion]{Figure~\ref*{fig:deepseek_confusion}} shows DeepSeek’s confusion matrices at LoRA ranks 4, 8, and 16. At $r=4$, DeepSeek achieves very high AI‐class recall but lower precision, with some false positives on human texts. At $r=8$, recall improves further at the expense of precision, indicating over‐sensitive AI detection. At $r=16$, both precision and recall recover, yielding the best overall trade‐off and minimal confusion.

\begin{figure}[ht]
  \centering
  \begin{subfigure}[b]{0.15\textwidth}
    \includegraphics[width=\textwidth]{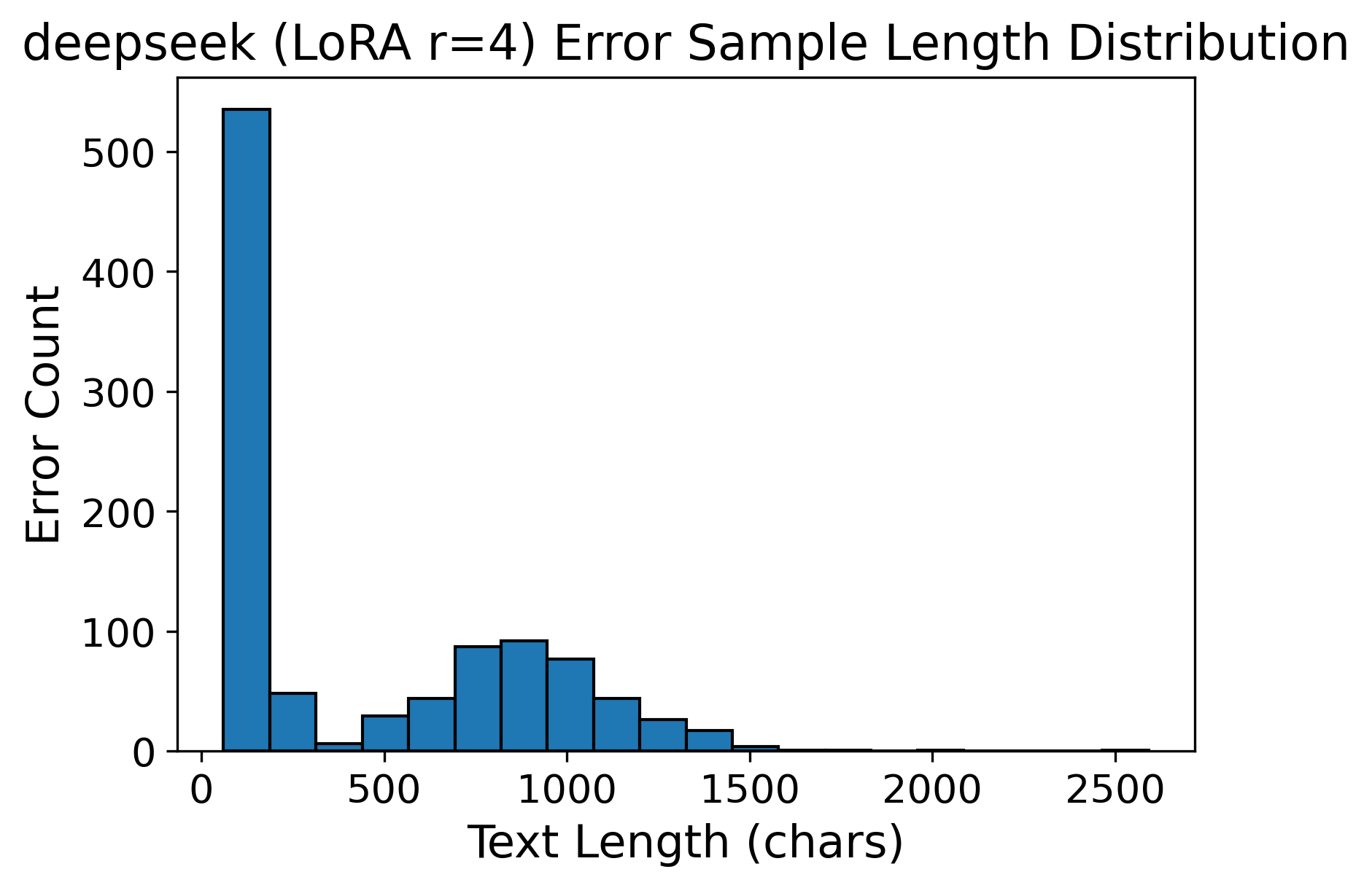}
    \caption{$r=4$}
    \label{fig:deepseek_r4_err}
  \end{subfigure}%
  \begin{subfigure}[b]{0.15\textwidth}
    \includegraphics[width=\textwidth]{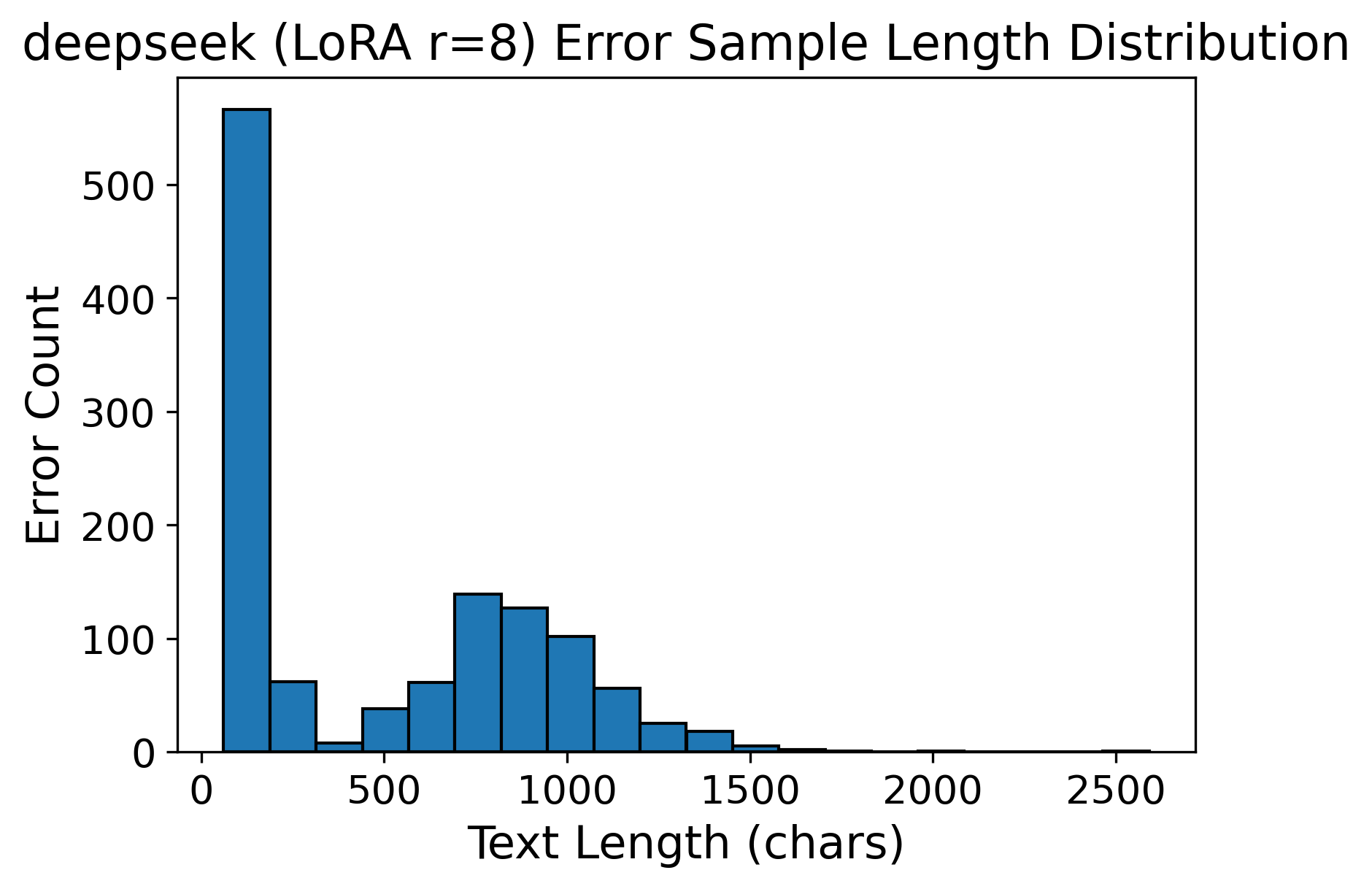}
    \caption{$r=8$}
    \label{fig:deepseek_r8_err}
  \end{subfigure}%
  \begin{subfigure}[b]{0.15\textwidth}
    \includegraphics[width=\textwidth]{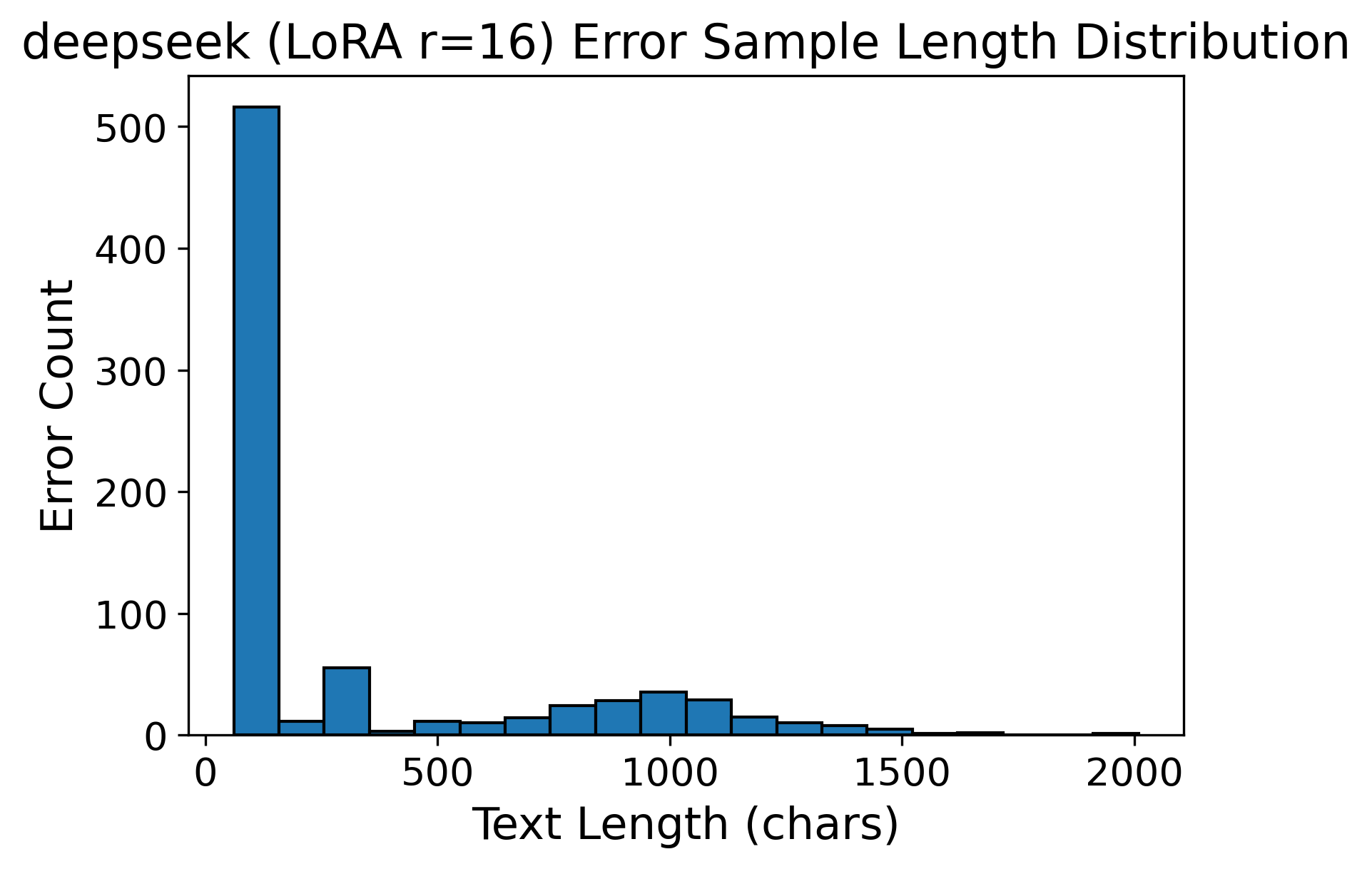}
    \caption{$r=16$}
    \label{fig:deepseek_r16_err}
  \end{subfigure}
  \caption{DeepSeek error‐length distributions for $r=4,8,16$.}
  \label{fig:deepseek_error_len}
\end{figure}

\hyperref[fig:deepseek_error_len]{Figure~\ref*{fig:deepseek_error_len}} demonstrates DeepSeek’s error‐length distributions for ranks 4, 8, and 16. At $r=4$, errors concentrate on medium‐length samples, reflecting limited capacity for longer patterns. At $r=8$, errors become more uniform across lengths, suggesting over‐generalization. At $r=16$, error counts drop sharply for all lengths, especially longer inputs, indicating robust, length‐aware performance.

\begin{figure}[ht]
  \centering
  \begin{subfigure}[b]{0.15\textwidth}
    \includegraphics[width=\textwidth]{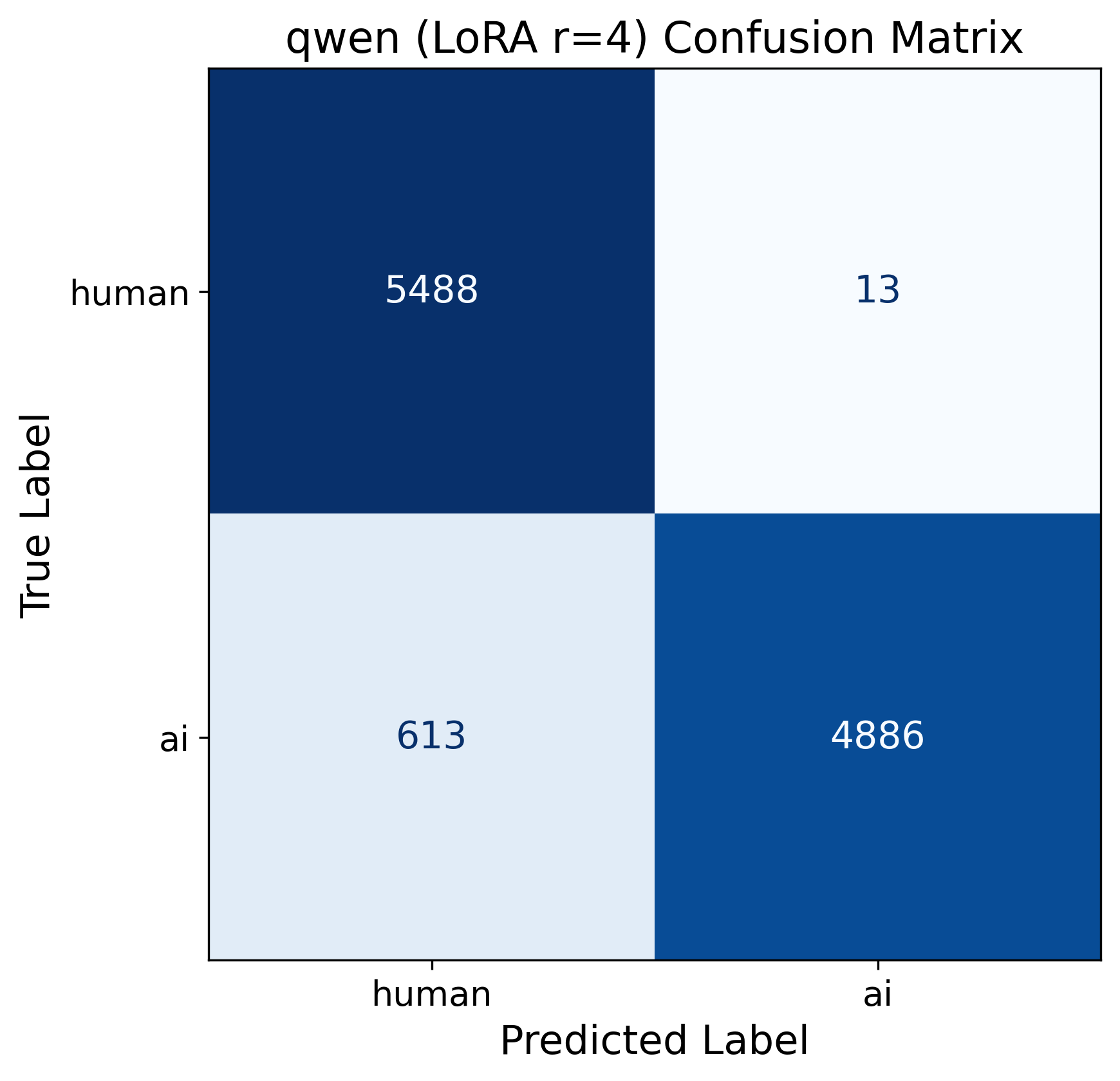}
    \caption{$r=4$}
    \label{fig:qwen_r4_conf}
  \end{subfigure}%
  \begin{subfigure}[b]{0.15\textwidth}
    \includegraphics[width=\textwidth]{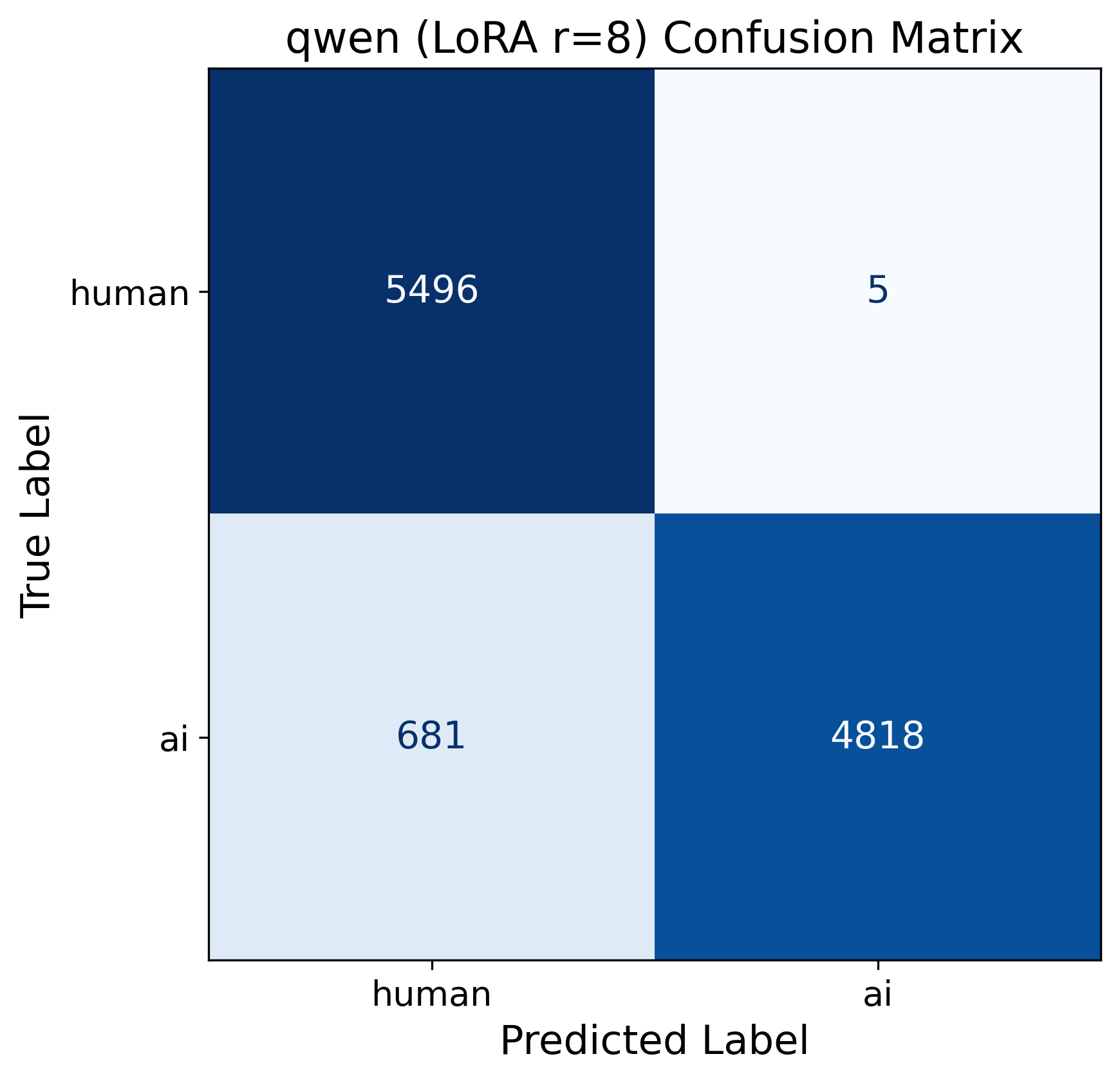}
    \caption{$r=8$}
    \label{fig:qwen_r8_conf}
  \end{subfigure}%
  \begin{subfigure}[b]{0.15\textwidth}
    \includegraphics[width=\textwidth]{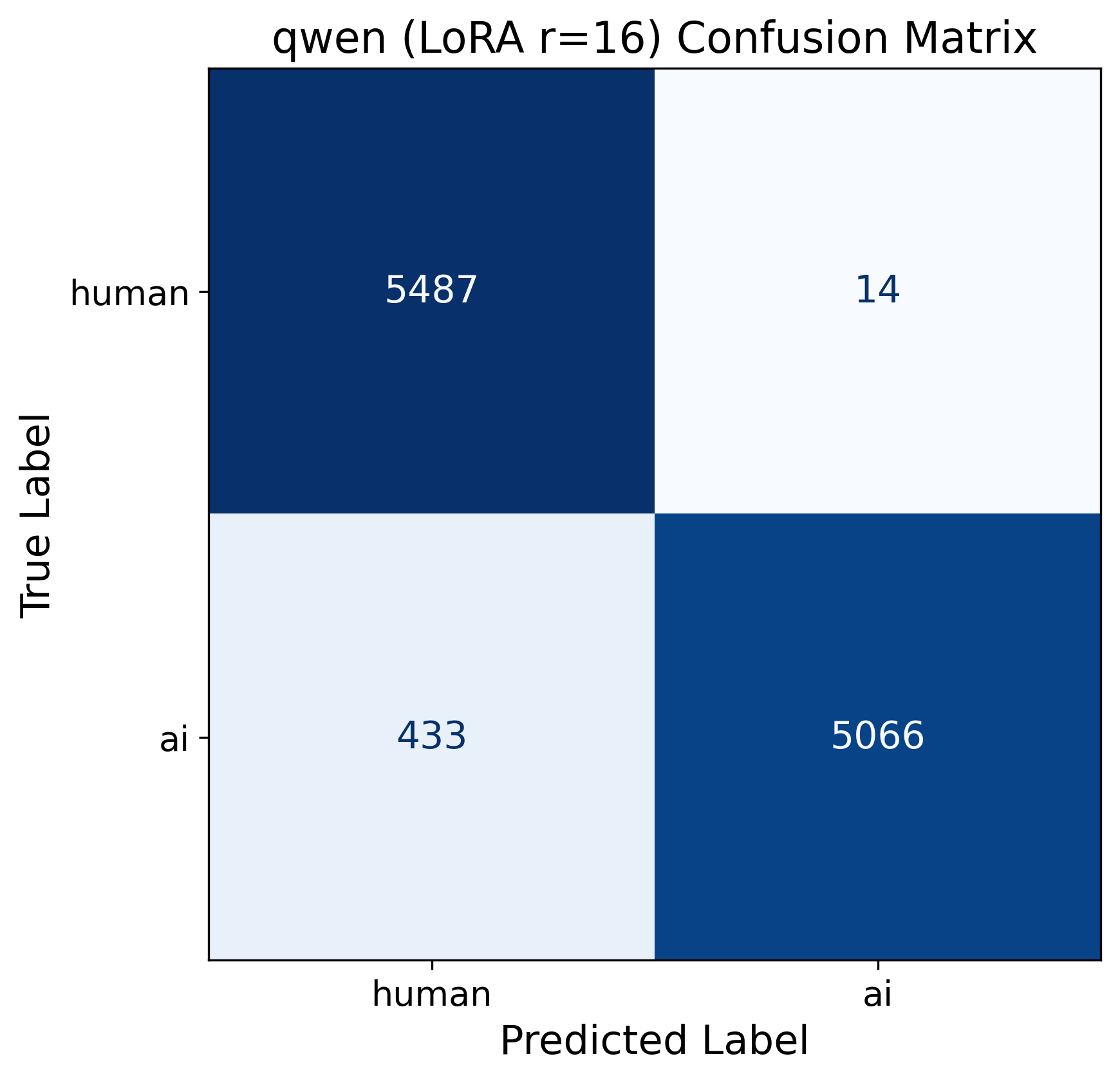}
    \caption{$r=16$}
    \label{fig:qwen_r16_conf}
  \end{subfigure}
  \caption{Qwen confusion matrices for $r=4,8,16$.}
  \label{fig:qwen_confusion}
\end{figure}

Qwen’s confusion matrices at LoRA ranks 4, 8, and 16 are shown in \hyperref[fig:qwen_confusion]{Figure~\ref*{fig:qwen_confusion}}. Even at $r=4$, Qwen balances precision and recall with minimal misclassifications. At $r=8$, AI detection becomes nearly perfect but slightly more conservative on human texts. At $r=16$, Qwen attains the optimal balance, minimizing both misclassified errors.

\begin{figure}[ht]
  \centering
  \begin{subfigure}[b]{0.15\textwidth}
    \includegraphics[width=\textwidth]{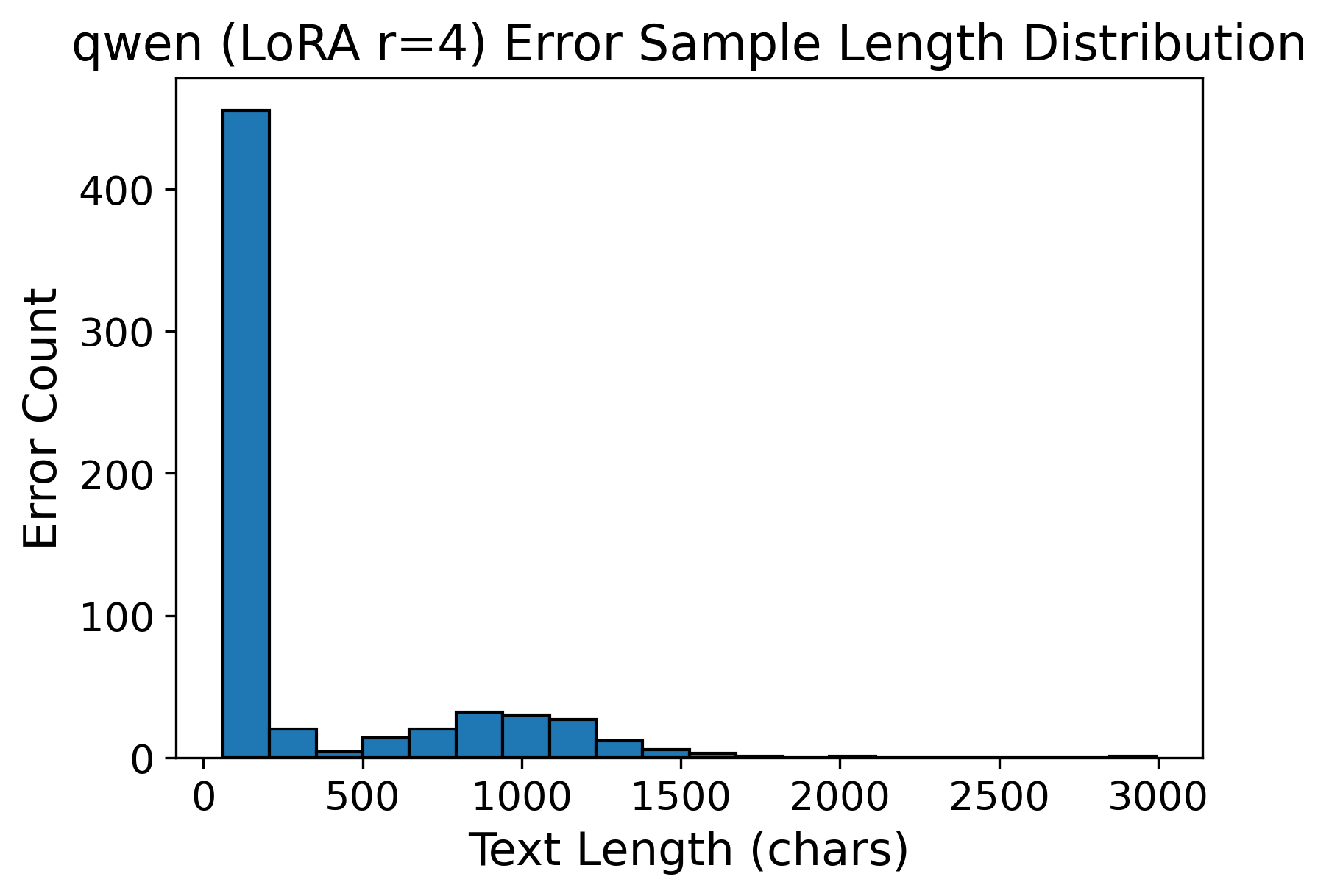}
    \caption{$r=4$}
    \label{fig:qwen_r4_err}
  \end{subfigure}%
  \begin{subfigure}[b]{0.15\textwidth}
    \includegraphics[width=\textwidth]{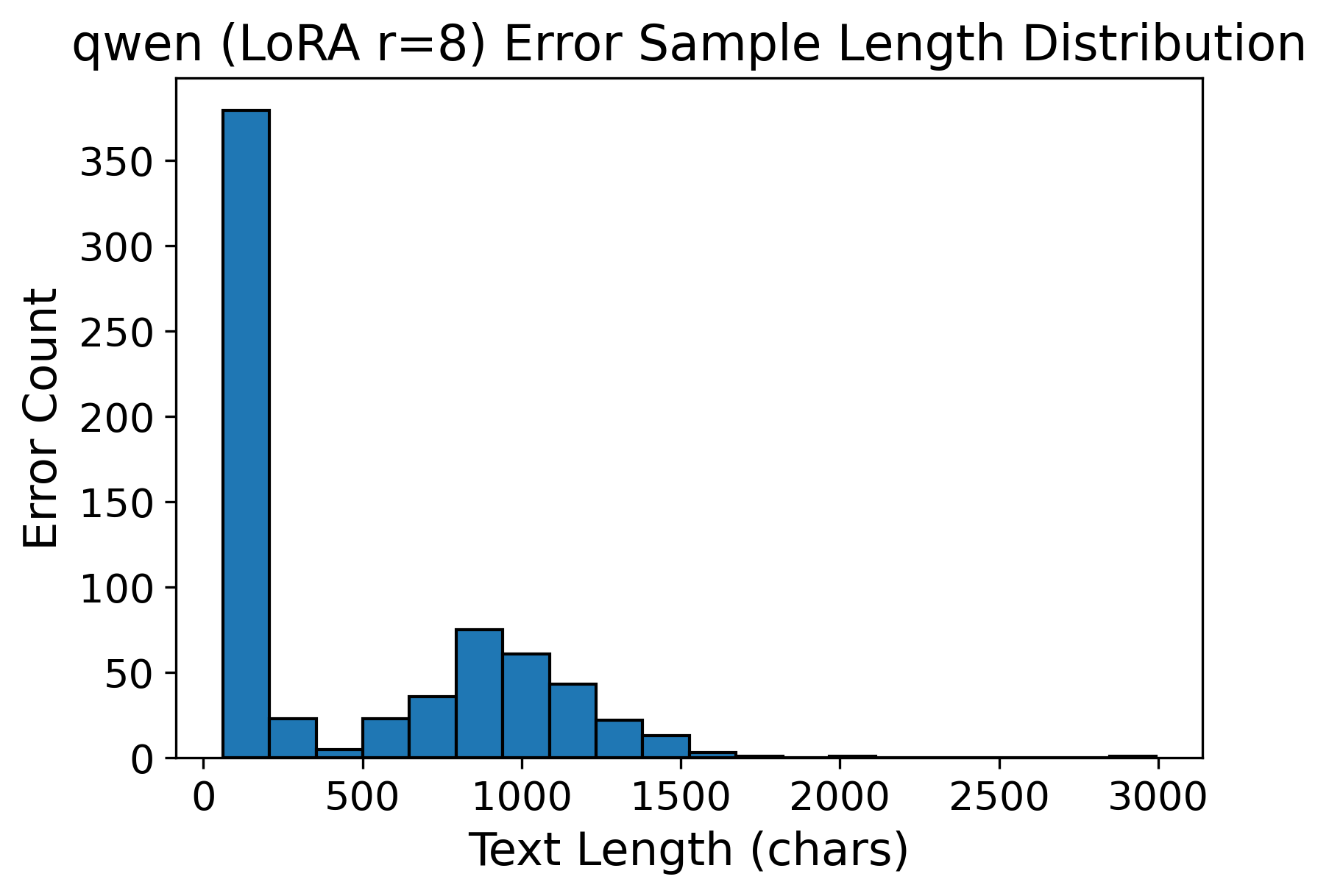}
    \caption{$r=8$}
    \label{fig:qwen_r8_err}
  \end{subfigure}%
  \begin{subfigure}[b]{0.15\textwidth}
    \includegraphics[width=\textwidth]{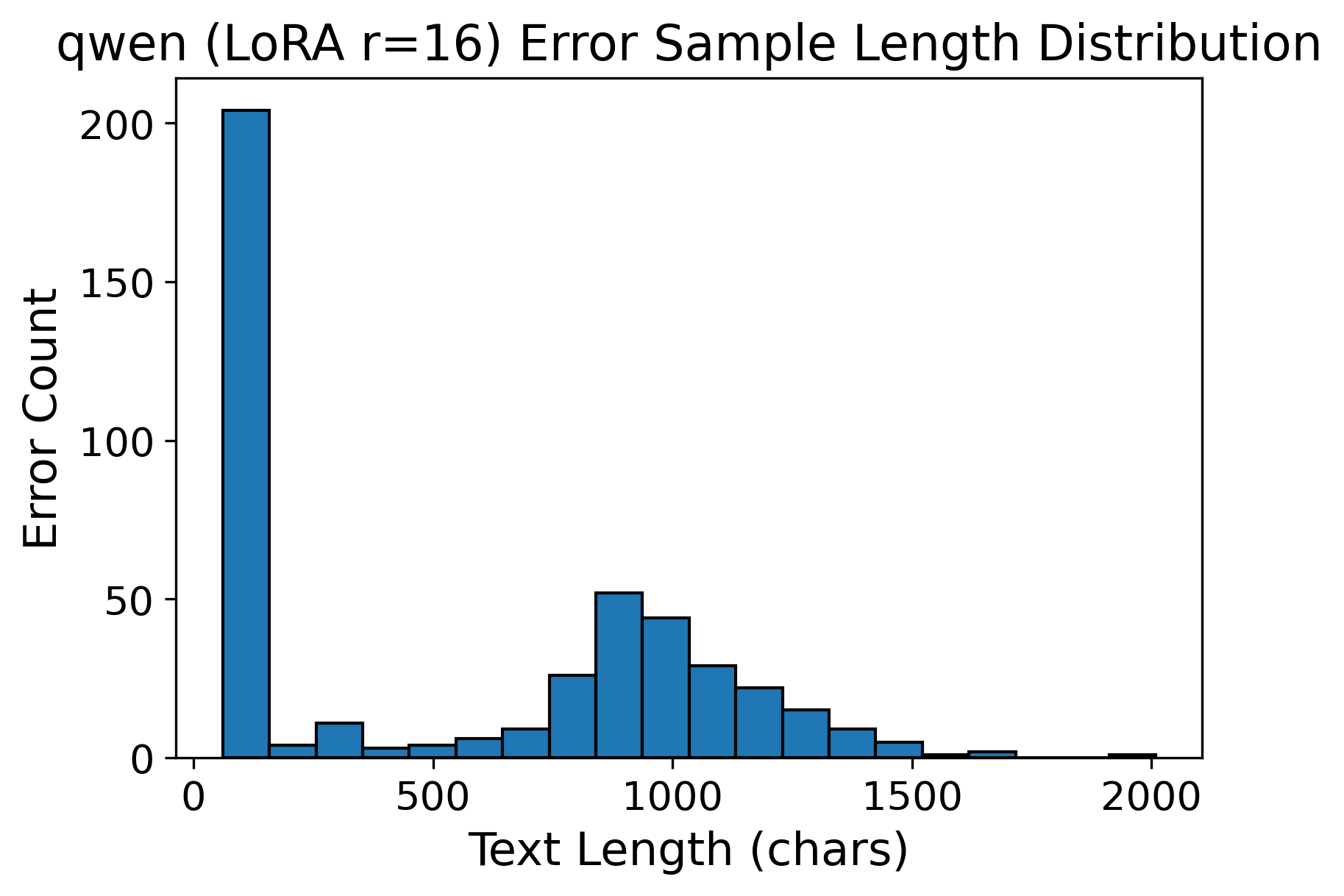}
    \caption{$r=16$}
    \label{fig:qwen_r16_err}
  \end{subfigure}
  \caption{Qwen error‐length distributions for $r=4,8,16$.}
  \label{fig:qwen_error_len}
\end{figure}

Qwen’s error‐length distributions at ranks 4, 8, and 16 are demonstrated in \hyperref[fig:qwen_error_len]{Figure~\ref*{fig:qwen_error_len}}. Across all ranks, Qwen maintains uniformly low error rates across text lengths, with slight improvements at higher ranks, reflecting its strong, generative pre‐training.

Initially, we reported results for Qwen2.5-7B with rank = 8. We later extended the evaluation to multiple LoRA rank settings for both Qwen (including rank = 8) and DeepSeek models, while keeping the original training logic unchanged. Minor differences in performance may arise due to variations in random seed initialization. Overall, Qwen consistently outperforms DeepSeek at every LoRA rank, achieving higher accuracy, tighter confusion patterns, and lower error rates across input lengths, whereas DeepSeek approaches this level of robustness only at its highest rank ($r=16$).

\section{Conclusion and Future Work}
\label{sec:conc}
This comparative study reveals striking architectural differences in Chinese AI-generated text detection performance. Qwen2.5-7B with LoRA fine-tuning achieves 95.94\% test accuracy, substantially outperforming encoder models that suffer catastrophic degradation under distribution shift (RoBERTa: 99.65\% $\rightarrow$ 76.31\%; BERT: 98.72\% $\rightarrow$ 79.33\%). The decoder's superior generalization stems from its large-scale generative pretraining on 2.4 trillion tokens, which appears to capture more transferable linguistic representations than the encoders' bidirectional but smaller-scale pretraining. The FastText baseline's unexpected robustness (83.50\% test accuracy) challenges assumptions about model complexity requirements, suggesting that distinguishing human from AI-generated Chinese text may rely heavily on surface-level lexical patterns for current generation models.

Our hypothesis that non-instruction-tuned models better capture AI text signatures requires cautious interpretation. While Qwen2.5-7B consistently outperformed the reasoning-distilled DeepSeek-R1-7B across all LoRA ranks, multiple confounding factors, including different distillation procedures, training objectives, and random initialization, preclude definitive conclusions. The performance gap may reflect architectural advantages rather than instruction-tuning effects. Additionally, evaluation on a single dataset limits generalizability claims, and our prompt-based masked language modeling approach for encoders lacks systematic comparison with standard classification heads.

Future research should prioritize cross-domain robustness evaluation using multiple Chinese AI detection datasets to validate architectural superiority claims. Ensemble methods combining FastText's lexical sensitivity with Qwen's semantic understanding represent a promising direction, potentially achieving both efficiency and accuracy. Systematic analysis of what linguistic features each architecture captures, through attention visualization, feature attribution, or adversarial perturbation, would illuminate the mechanistic basis for their performance differences and guide architecture selection for specific deployment contexts.

\bibliography{references}

\end{CJK*}
\end{document}